 \documentclass[final,3p,times,twocolumn]{elsarticle}
\usepackage{amssymb}
\usepackage{graphicx}
\usepackage{algorithm}
\usepackage{algorithmicx}
\usepackage[noend]{algpseudocode}
\usepackage{subfigure}
\usepackage{epsfig}
\usepackage{amsmath}
\usepackage{booktabs}
\begin{document}

\begin{frontmatter}


\title{Metric Map Merging using RFID Tags \& Topological Information}
 \author{E. Tsardoulias}
 \ead{etsardou@eng.auth.gr}
 \author{A. Thallas}
 \ead{athallas@ece.auth.gr}
 \author{L. Petrou}
 \ead{loukas@eng.auth.gr}


 \address{
 Aristotle University of Thessaloniki,\\
  Faculty of Engineering,\\
  Department of Electrical and Computer Engineering,\\
  Division of Electronics and Computer Engineering,\\
  54124 Thessaloniki, GREECE \\
  Tel.: +302310996294\\
  Fax: +302310996447
 }

\begin{abstract}
A map merging component is crucial for the proper functionality of a
multi-robot system performing exploration, since it provides the means to integrate and
distribute the most important information carried by the agents: the
explored/covered space and its exact (depending on the SLAM accuracy)
morphology.
Map merging is a prerequisite for an intelligent multi-robot team
aiming to deploy a smart exploration technique.
In the current work, a metric map merging approach based on environmental
information is proposed, in conjunction with spatially scattered RFID tags
localization.
This approach is divided into the following parts: the maps' approximate
rotation calculation via the obstacles' poses and localized RFID tags,
the translation employing the best localized common RFID tag
and finally the transformation refinement using an ICP algorithm.

\end{abstract}


\end{frontmatter}


\section{Introduction}
\label{intro}

One of the most promising aspects of rescue robotics is the employment of multiple
robots to simultaneously operate in a disaster afflicted environment.
A robotic team's deployment presents significant benefits compared to
single agent approaches.
First of all, the overall performance is boosted as larger areas
are explored and covered per time unit, due to the
robots' distributed operation.
Next, a multi-agent system is more robust, as the exploration may resume
even after a potential robot failure.
Finally, the deployment of many identical or heterogeneous robots vastly reduces
such a system's cost, as a single robot equipped with
expensive sensors can be replaced with an ensemble of cheap vehicles with
limited sensing capabilities.

In order for multiple robots to achieve efficient cooperation,
a mechanism should exist to support the sharing of
their distributed information.
Even if they are able to flawlessly communicate with each other,
it is impossible to properly
merge specific parts of information, i.e. map representations, without the
existence of a common reference.
In the simplest case, \emph{common reference} refers to a
two dimensional transformation
between the robotic agents' local coordinate systems.
Thus, it is understandable that the aforementioned transformation should be
efficiently calculated in order for true group intelligence to exist.

In the current work, the map representation used is OGM (occupancy grid map).
OGMs are metric depictions of the environment, consisting of a grid of cells,
each of which is assigned an occupancy probability value.
A specific cell's occupancy probability is altered according to the current
range sensors' measurements.
Our main focus is to create an efficient approach towards merging local OGMs
using topological information and RFID tags, by dividing the problem into the
calculation of the translation and rotation separately.
Our proposal consists of several consecutive steps.
The first step is to accurately compute an alignment angle concerning the
environmental obstacles, which we assume to be orthogonal.
Next, the correct rotation quaternion is determined via the topological alignment
of common RFID tags' poses among the robots.
The third step implements the translation calculation employing the best
localized pair of RFID tags.
These steps can achieve a quite good approximation of the actual transformation.
However, in order to achieve an exact alignment, an additional step is introduced,
including the deployment of an ICP (Iterative Closest Point) algorithm.
Finally, due to the discrete map representation, inconsistencies appear after
the transformation's application.
This issue is resolved using a modified blurring process.

The paper is organized as follows.
In section~\ref{sec:sota} the State of the Art concerning the map merging problem
is presented.
Section~\ref{sec:rfid_tag_localization} describes the RFID tag localization
procedure and section~\ref{sec:metric_map_merging} contains the actual steps
of the map merging implementation.
In section~\ref{sec:experiments} the experiments are presented
and finally in section~\ref{sec:conclusions} conclusions and future work
are discussed.

\section{Related Work}
\label{sec:sota}

In \cite{carpin2005stochastic}, Carpin and Birk suggest a quite straightforward
map merging method.
The proposed solution implements a stochastic search in the
map transformations space, comprised of translations and rotations.
This search method consists of a time variant Gaussian Random
Walk, altering its distribution parameters in order to employ
its recent values' history, approaching the correct solution.

A different method is presented in \cite{sun2009multi}, where the data association
problem is resolved using wireless sensors' IDs, detectable
by an omnidirectional 2.4 GHz antenna.
These are employed for a map merging common reference frame calculation.
Each robot submits its pose in the wireless sensor memory,
allowing robots to acknowledge that they share a common explored area.
A similar approach is followed in the current work, suggesting the use of
spatially scattered RFID tags, whose pose is probabilistically localized
and, if special conditions are met, map merging is performed.
In our approach we experimentally prove that without the exact
wireless sensors' pose it is impossible to achieve precise map merging, thus
extra optimization methods are required.
The original probabilistic technique for the RFID
tags' pose calculation using Bayes filters is presented in
\cite{hahnel2004mapping}.
A two dimensional probability distribution of a single tag is
calculated, based on the actual RFID antenna and
reader operating field, whilst in our case an ideal uniform omnidirectional
antenna was assumed.

In \cite{konolige2003map}, Konolige, Fox et al., suggest a distributed mapping
technique under uncertain communication, as well as an algorithm
for the robot poses localization in the local maps.
Specifically, the $i^{th}$ robot's pose in the $j^{th}$ robot's local map,
is calculated by using a small fraction of $i$'s map, in which three distinct
topological features are recognized: corners, doors and junctions.
Next, following a similar procedure in $j$'s map and by utilizing probabilistic
methods, the dominant $i$'s pose is calculated, enabling the information
exchange.

Following a similar approach incorporating features, Amigoni and
Gasparini suggest a map merging method with no  odometry
information requirements in \cite{simmons2000coordination}.
This allows for map merging just from the map representations,
providing flexibility to the algorithm.
In this publication, the features are corners resulting from the straight line
segments which form the environmental obstacles.
The merging procedure consists of three steps:
a) the creation of possible transformations based on feature matching,
b) the best transformation identification and
c) the transformation application in the robots' local maps.

Furthermore, in \cite{bonanni2014} the merging of partially consistent maps
is considered.
There, the merging problem is reduced to the problem of deforming two networks
(or graphs) onto each other in order to minimize the ``residual energy''.
Obviously, this approach requires the existence of a pose graph, which in the case
of an OGM is acquired via the unoccupied space's Voronoi diagram.
Additionally, in \cite{saeedi2014} the \textit{Group Mapping} approach is
proposed.
There, the generalized Voronoi diagram (GVD) is extended to include the
probabilistic information stored in the OGM, which result is denoted as
PGVD (Probabilistic Generalized Voronoi Diagram).
This is then used to determine the maps' relative transformation and is
utilized for their merge.
Finally, in \cite{lazaro2013} each robot refines its map by integrating
a set of ``condensed measurements'' deriving from the other agents.
This way a small amount of information is utilized and communicated,
instead of the whole map.
Then, a RANSAC algorithm is deployed to perform data association in order
to localize one robot in another robot's condensed measurements graph.

Moving on from the feature-based merging techniques to iterative algorithms,
a different approach is proposed in
\cite{fox2006distributed}.
The robots exchange some of their scans in time slots and perform
localization techniques in their local maps, using Rao-Blackwellized particle
filtering.
Additionally, concurrently to the SLAM execution, a graph-based
restrictions system is maintained.
These restrictions are produced by the laser scans.
When the localization system calculates a robot's pose in
another robot's local map, these restriction rules are utilized to
accurately align the two information sources.

In \cite{carpin2005map}, Carpin, Birk and Jucikas suggest an iterative procedure in
order to compute the appropriate transformation between two local maps.
Specifically, a dissimilarity function is defined,
indicating the extend of dissimilarity between the two maps.
Initially, a geometric transformation is suggested and its dissimilarity
function value is calculated.
Then, the ``Random Walk'' algorithm is executed for each algorithmic iteration,
where the current transformation is slightly mutated, creating an ensemble of
others, for whom the dissimilarity value is calculated too.
Next, the transformation corresponding to the lower dissimilarity value is
selected.
The algorithm's utmost goal is to minimize the dissimilarity function.
This algorithm highly resembles genetic procedures or hill climbing
algorithms, where a fitness function minimization (or maximization)
effort is performed.

Similarly to the dissimilarity function, in \cite{li2014}
Li et. al. propose an objective function based on occupancy likelihood,
which is optimized via genetic-like procedures.
Furthermore, a strategy of vehicle-to-vehicle relative pose estimation is
provided, which serves as a general solution for multi-vehicle perception
accusation.

In \cite{carpin2008fast}, Carpin describes the construction of a mathematical,
non iterative method, performing fast and accurate merging of local
maps.
According to this, a geometric transformations set is created,
containing the possible merging solutions.
Then, a weight is calculated for each transformation, allowing to
identify uncertain situations and enabling the tracking of multiple hypotheses
when ambiguities are present.
These transformations are computed via the Hough spectral information
and are utilized in the calculation of the relative maps' rotation.
Similarly, the translation through Hough spectral information is calculated,
thus completing the transformation solution.
This approach poses similarities with \cite{saeedi2012}, where the individual
maps are transformed in the Hough space.
Then the correct transformation is calculated via identification of common
areas among the maps, based on the Hough space properties.
The rotation calculation methodology is similar to the one performed in the
current work, with the difference that here an enhanced RANSAC algorithm is
proposed instead of Hough.

Finally, in \cite{leung2011utias}, a collection of 2D datasets, oriented in
multi-vehicle map merging is presented, called \emph{UTIAS}.
This enables for multi-robot pose calculation, localization, mapping and
other robotic problems research.

\section{RFID Tag Localization}
\label{sec:rfid_tag_localization}

The following assumptions are made regarding the RFID system's behavior.
Firstly, we assume that scattered RFID tags can
be identified by omnidirectional antennas placed in every vehicle.
The nominal detection range of each antenna is $3~m$.
Additionally, an RFID antenna receives three different types of information:
the global RFID tag's ID, its internal memory contents and the wireless signal's
strength in dBs.
It is obvious, that the sole way of calculating the tag's distance from the antenna
resides in the signal strength.
However, we assume that a module exists, providing the RFID tag
distance from the antenna, by processing the signal's dBs.
Finally, the distance measurements are susceptible to noise, which assumingly
follows a Gaussian distribution with zero mean and standard deviation equal to
$\sigma = 5~cm$.

\begin{figure}[htbp]
  \centering
  \includegraphics[width=0.8\linewidth]{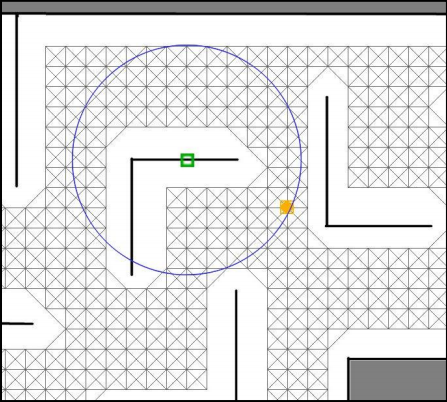}
  \caption{Conditions for the RFID tag localization procedure initialization}
  \label{fig:rfid_loc_con}
\end{figure}

In figure~\ref{fig:rfid_loc_con}, the RFID tag localization procedure initialization
conditions are presented.
An RFID tag, placed on an obstacle surface, is depicted with green
color and the blue circle denotes the RFID antenna's maximum detection range.
The robot (where the RFID antenna is placed) is the orange colored cube.

The presented technique for probabilistically localizing an RFID tag,
is based on a two dimensional probability function $f(x,y)$,
updated in each algorithmic iteration, i.e. every time the algorithm
receives new information about the surrounding RFID tags.

\begin{figure}[htbp]
  \centering
  \includegraphics[width=.8\linewidth]{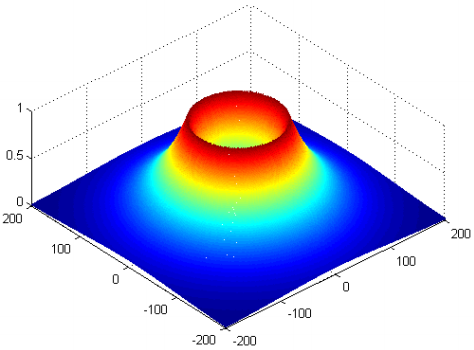}
  \caption{RFID tag pose probability distribution (axis in pixels)}
  \label{fig:exa_prob_fun_rfid}
\end{figure}

During an RFID tag localization with $I\!D=i$, a circularly shaped, two
dimensional probability density function $f_t$ is declared,
having as center the current robot pose $P_R$ and radius equal to
the distance $D$ from the RFID tag to the robot.
This probability distribution has its maximum on the circumference of
the circle with radius $D$, decreasing smoothly in the surrounding area.
An example of the probability function is presented in
figure~\ref{fig:exa_prob_fun_rfid}.

\begin{figure}[htbp]
  \centering
  \includegraphics[width=.8\linewidth]{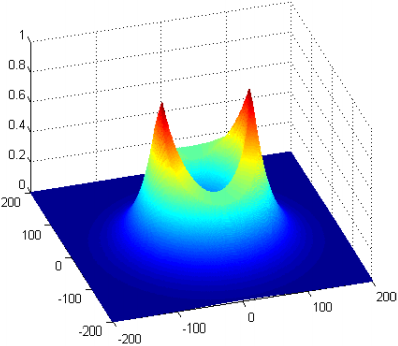}
  \caption{Product of two circular shaped probability distributions}
  \label{fig:mult_prob}
\end{figure}

The next step is to incorporate this distribution in the total probability
distribution $f_{total}$, from which the most probable tag pose will be
determined.
Let's assume that in the next data update the same tag is perceived at a
distance equal to $D'$.
Of course, the robot has moved and its current pose is $P_R'$.
Again, a temporary probability distribution $f_t$,
centered around $P_R'$ with a radius of $D'$ is created.
Next, the total probability distribution is updated by multiplying its
current values with the new information, i.e. $f_{total} = f_{total} \cdot f_t$.
This formula constitutes another form of the Bayes probabilistic filter,
described by:

\begin{align}
  \label{eq:rfid_bayes}
  p( x^{T}_{1:t} \,|\, z_{i:t}, x^{R}_{1:t} ) =
    \alpha \cdot &p( x^{T}_{t} \,|\, x^{T}_{t-1}, x^{R}_{t}) \nonumber \\
           \cdot &p( x^{T}_{1:t-1} \,|\, z_{1:t-1}, x^{R}_{1:t-1})
\end{align}

An example of two probability distributions multiplication result
is visible in figure~\ref{fig:mult_prob}.
Its shape -- specifically the two maxima -- is expected, since the
initial distributions were circularly shaped and the maxima exist on the
circles' cross-sections.

\begin{figure}[htbp]
  \centering
  \subfigure[Probability distribution after 30 iterations]
  {
    \centering
    \includegraphics[width=0.45\linewidth]{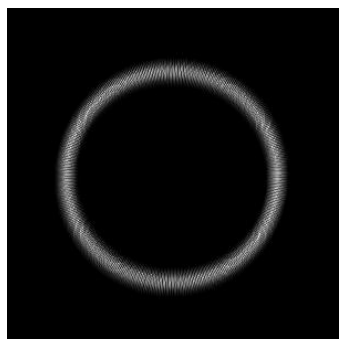}
    \label{fig:rfid_30}
  }
  \hspace{2mm}
  \subfigure[Probability distribution after 100 iterations]
  {
    \centering
    \includegraphics[width=0.45\linewidth]{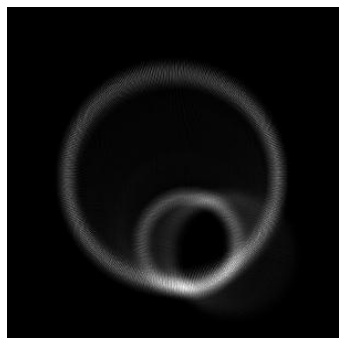}
    \label{fig:rfid_100}
  }

  \vspace{2mm}

  \subfigure[Probability distribution after 200 iterations]
  {
    \centering
    \includegraphics[width=0.45\linewidth]{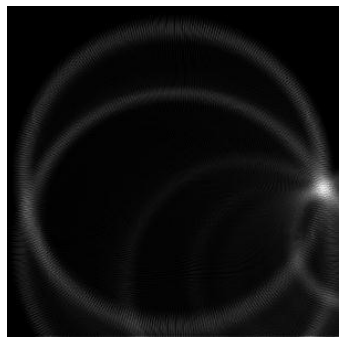}
    \label{fig:rfid_200}
  }
  \hspace{2mm}
  \subfigure[Probability distribution after 300 iterations]
  {
    \centering
    \includegraphics[width=0.45\linewidth]{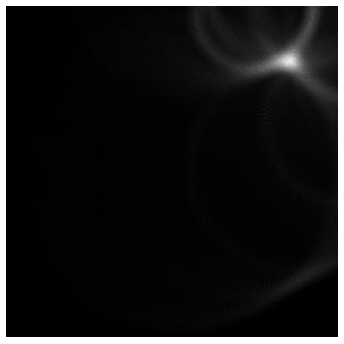}
    \label{fig:rfid_300}
  }
  \caption{Global probability distributions for a specific RFID tag for 30, 100,
    200 and 300 iterations}
\end{figure}

This procedure is performed until the specific tag is not perceived by the robot,
i.e. the distance between the robot and the tag exceeds the $3~m$ threshold.
The localization certainty of a tag's pose is proportional to the
time perceived by the robot, measured in iterations.
In figures \ref{fig:rfid_30} to \ref{fig:rfid_300} the global
probability distribution for a specific RFID tag and for 30, 100, 200 and 300
iterations is depicted.

After the final probability distribution's calculation, the
identification of the most probable tag pose is performed, as well as
the specification of a way to determine whether the pose calculation is accurate.
A reasonable selection would be the cell for which $f_{total}$ is
maximized, which however, results in a complication in the correct
pose's probability identification.
Since the distribution is not continuous (every cell holds a distinct
value), the probability corresponding to a cell, is the division of
its value to the sum of all the other cells, i.e.:

\begin{equation}
  p_{x,y} =
  \frac{f_{total}(x,y)}
  { \sum_{ \forall x_i, y_j \in f_{total} } f_{total}(x_i, y_j)}
\end{equation}

The problem is that due to the large number of cells
participating in the distribution,
$p_{x,y}$ is extremely small, even for cases where the
correct pose can be obviously identified.
Even though this problem is not severe, the drawback is mostly
aesthetical since the calculated certainty probability does not coincide with
human perception.
For example, a human would identify a $90\%$ localization probability as
pretty accurate, but in reality it is impossible such a value to occur.
Thus, we decided to extract the probability of the cell's surrounding area,
instead out of a single pose.
Specifically, if the initial probability distribution included $N\cdot M$ values,
a sub sampling by $K$ is performed in order to produce
$\frac{N}{3}\cdot\frac{M}{3}$ values (in our case $N = 300$, $M=300$ and $K=3$).
Each cell's value is the arithmetic mean of the cells comprising the $3 \times 3$
square area that surrounds it.
This way, the values' differences are ``smoothed'', the number of
values decreases and the actual probability calculated is quite realistic.
This procedure can be mathematically supported;
since the probability distribution is discrete but contains a large number
of values, it can be perceived as a continuous one.
As we know, in continuous distributions a point's probability
to contain a specific value is $0$.
To obtain a more numerically representative value, an integration must be
performed.

\section{Metric Map Merging}
\label{sec:metric_map_merging}

The current work addresses the issue of merging
a multi-robot team's distributed information, whose initial poses are a priori unknown.
The lack of knowledge regarding the starting positions
connotes that robots lack the required reference frame to properly
communicate and transform their information.
Furthermore, it is assumed that robots are incapable of physically
identifying each other, i.e. using cameras or distance sensors.
As a result, the problem to be resolved is to specify the correct
transformation -- rotation and translation -- between the robots' coordinate
systems, purely via common environmental information.

Our approach towards specifying the transformation is performed in three
consecutive steps, to be described in detail: a) calculation of an OGM direction
vector, b) calculation of the rotation quadrant and the relative translation
via RFID tag information and c) exact transformation specification employing
an ICP algorithm.

\begin{figure}[t]
  \centering
  \subfigure[First Map]
  {
    \centering
    \includegraphics[width=0.45\linewidth]{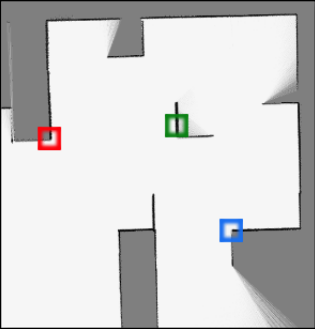}
    \label{fig:merge_example_1}
  }
  \hspace{2mm}
  \subfigure[Second Map]
  {
    \centering
    \includegraphics[width=0.45\linewidth]{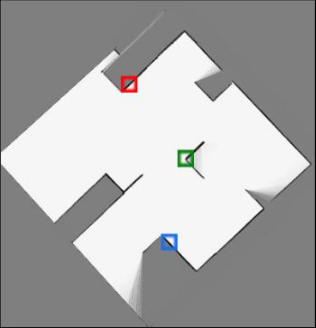}
    \label{fig:merge_example_2}
  }

  \vspace{2mm}

  \subfigure[Merge result]
  {
    \centering
    \includegraphics[width=0.9\linewidth]{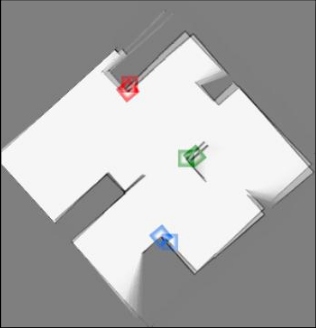}
    \label{fig:merge_example_3}
  }
  \caption{Example of approximate OGM merging via three common RFID tags}
  \label{fig:rfid_map_merge}
\end{figure}

It should be stated that
scattered RFID tags exist in the environment where the robots operate.
As described in section~\ref{sec:rfid_tag_localization}, each agent can
localize RFID tags using a two dimensional probability distribution.
From this distribution the tag's pose can be extracted, as well as the
localization certainty.
Since each RFID tag has a unique identification number, it is possible to
calculate a rough transformation estimation between two coordinate
systems, provided that the two robots have already localized at least three common
RFID tags.

In figure~\ref{fig:rfid_map_merge} such a situation is presented.
The first two images contain the common RFID tags in the two environments,
whilst the third one presents an attempt to merge the maps, based on the RFID poses.
The calculation of the exact transformation between
two robots' coordinate systems presupposes the exact RFID tags' pose
localization, something impossible, since this is performed
probabilistically and not in an analytical way.
Thus, the produced transformation will be approximate, since
false occupancy values could exist after the map merging.

The transformation computation between
two local maps is initiated when the two robots have localized at least three
common RFID tags, whose localization probabilities exceed a predefined threshold.
In the current work, this minimum probability value was selected equal to $75\%$.
Additionally, for reasons to be analyzed further below, there is a prerequisite
that one of the three RFID pairs has a quite high localization probability,
whose threshold was set to $90\%$.

\subsection{Relative Rotation Calculation via OGM Vectors}
\label{sub:rotation_calculation_vector}

\begin{figure}
  \centering
  \includegraphics[width=\linewidth]{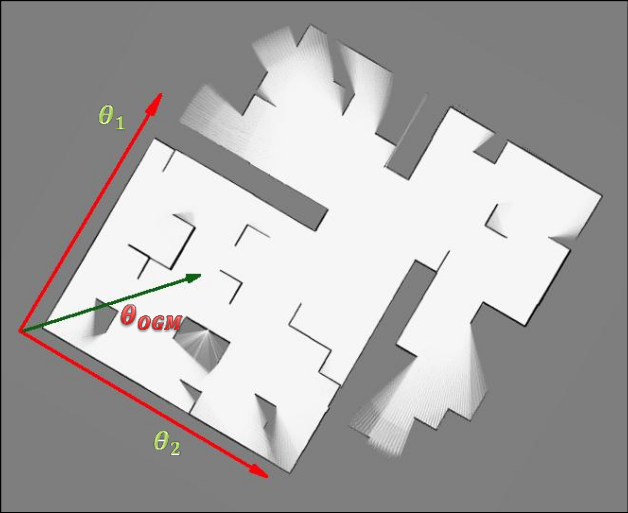}
  \caption{OGM direction vector}
  \label{fig:perpend_obst}
\end{figure}

\emph{OGM Vector} constitutes a feature originating from the environment in
conjunction with an assumption.
Specifically, it is assumed that the robot explores a civil environment, having
a rectangular form, implying that the environmental obstacles are
orthogonal to each other.
An example is shown in figure \ref{fig:perpend_obst}.

In this figure the two red arrows represent the two perpendicular directions
$\theta_1, \theta_2 \, (\theta_1 \!\! \perp \!\! \theta_2)$,
 derived from the obstacles, whilst the green arrow specifies the OGM
direction vector, whose orientation is equal to
$\theta_{OGM}=\frac{\theta_1 + \theta_2}{2}$.
It should be mentioned that angles $\theta_1, \theta_2$ should be
bounded $\left[-\frac{\pi}{2}, \frac{\pi}{2} \right]$, which results in
$\theta_{OGM} \in \left[ -\frac{\pi}{2}, \frac{\pi}{2} \right]$.
The concept behind the computation of OGM's direction vector is the following:
let's assume that the two local OGMs are available and their relative
transformation is unknown, preventing the merge.
Since the robots physically exist in the same space, by determining the
direction vectors $\theta_{OGM_1}, \theta_{OGM_2}$ it is possible to calculate the
angle $\Delta\theta_{OGM}=\theta_{OGM_1}-\theta_{OGM_2}$, which
if applied to one of the maps aligns its obstacles to the other's.
It should be noted that this procedure aims at calculating a high precision
approximation of the alignment angle, but not the final transformation's rotation.
This is due to the fact that the obstacles can be aligned in four
different ways, one for every rotation by $\frac{\pi}{2}$.
Thus, the real transformation rotation between the two maps is one of the
following:
\[
  \Delta\theta_{OGM} =
  \left\{
    \begin{array}{c}
      \theta_D \\
      \theta_D +\frac{\pi}{2} \\
      \theta_D+\pi \\
      \theta_D+\frac{3\pi}{2}
    \end{array}
  \right.
\]
The method to specify the correct alternative out of the four possible ones,
will be described in subsection~\ref{sub:translation_calculation_rfid}.

\subsubsection{RANSAC Line Detection}
\label{ssub:ransac_line_detection}

In order to calculate the OGM Vector, the first step is to detect the
straight segments corresponding to the map's obstacles.
One of the most common methods for this purpose is the RANSAC (RAndom SAmple
Consensus) algorithm.
Its application is repetitive and works as such:

\begin{enumerate}
  \item{
      A point set $P$ is created, from which we desire to extract line segments.
      In our problem, $P$ contains a sub sampling of the occupied OGM cells.
      The sub sampling itself was performed in order to increase the method's
      execution speed.
      In the current approach $N_{init}=|P|$.
    }
  \item{
      Two elements of $P$ are randomly selected ($p_1, p_2 \in P$).
    }
  \item{
      The line segment connecting $p_1, p_2$ is calculated
      and all $P$'s elements having a smaller distance than a
      predefined threshold $D_{Line}$ from the line, are inserted in the
      $L$ set.
    }
  \item{
      If $L$ contains a satisfactory percentage of $P$'s total points, i.e. if
      $\frac{|L|}{N_{init}}>T_{Perc}$, where $T_{perc}$ the percentage
      threshold, this line is registered as valid and $L$'s elements are erased
      from $P$.
      This procedure is quite flexible and can be implemented in many ways, one
      of which is to check $L$'s cardinality, instead of the percentage.
      If this condition does not apply, the algorithm returns in step 2.
    }
  \item{
      The procedure described in steps 2-4 is executed until a predefined condition
      seizes to be valid.
      A classic condition for RANSAC termination is $\frac{|P|}{N_{init}}<
      T_{final}$, i.e. the points percentage not registered in any
      lines, divided by the initial points number, to be sufficiently
      small.
    }
\end{enumerate}

The RANSAC algorithm is highly parameterizable, resulting in different behaviors
and execution times when different parameter sets are used.
For example, if the $T_{Perc}$ value is high (e.g. more than $30\%$)
and the environment is complex, the algorithm does not manage to detect any
lines (or a few will be detected with difficulty).
The same occurs if $T_{final} \approx 0$.
Nevertheless, if $T_{Perc}$ is assigned with a very small value, a large
number of lines will occur, which of course will contain an abundance of
false positive cases.
Aiming at minimizing the occurrence of these situations,
a limitation to the algorithm iterations number is usually applied.
Should the algorithm's iterations surpass this threshold,
the best so far computed line is maintained.
Finally, the increment of the $D_{Line}$ threshold,
implies greater tolerance of the algorithm with respect to the lines
which are not entirely aligned with the two initial random points.
On the contrary, if $D_{Line}$ is too small, the algorithm will not be able
to detect any lines, due to failure of the $T_{Perc}$ condition.

The former analysis dictates that a careful adjustment of the algorithmic
parameters is imperative, in order to achieve the goal and at the
same time keep the execution times low.
In our case, only a subset of the environment lines is needed,
under the condition that these lines are precise.
For this reason the $D_{Line}$ parameter was assigned a small value ($2~px$),
$T_{Perc}$ a relatively high value ($15\%$) and $T_{final}$ a low value
($50\%$).

Once a line $L_i$ is assumed valid, its new limits are calculated.
The new limits are the points which are members of the line and have the
maximum distance between them:
\[
\bigg\{ p_1', p_2' \in L_i \,|\, argmax_{p_1' p_2'} Dist(p_1',p_2') \bigg\}
\]

Thus, the real line segment's limits are calculated, since the initial points
$p_1, p_2$ are probably internal.
Theoretically, since the correct boundary points are calculated, the line's gradient
computed by those, is more accurate than the initial assumption.
For this reason, set $P$ is recalculated containing only the elements
having a maximum distance of $D_{Line}$ from the updated line.
For completeness reasons, the equation to calculate a point $A$'s distance
from a line, defined by two other points $B$ and $C$, follows:

\begin{equation}
  D_L = \frac{ |(x_C-x_B)\cdot(y_B-y_A)-(x_B-x_A)\cdot(y_C-y_B)| }
  {\sqrt{(x_C-x_B)^2 + (y_C-y_B)^2} }
\end{equation}

\begin{figure}
  \centering
  \includegraphics[width=\linewidth]{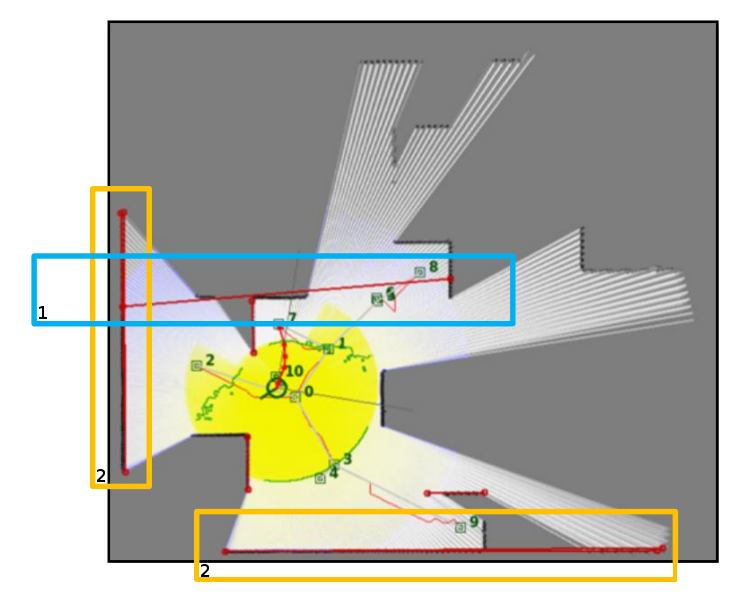}
  \caption{RANSAC-based line detection}
  \label{fig:ransac1}
\end{figure}

An example of the algorithmic execution is depicted in figure
\ref{fig:ransac1}.
The first observation is that the produced lines are not representative of the
overall obstacles of the environment, something expected since $T_{final}$ had a
low value.
Additionally, two undesirable results are detected.
The first erroneous result is the line contained in the blue box (labeled with
$1$), which is clearly not associated with an obstacle, but its points
lie in multiple occupied surfaces.
Furthermore, as evident in the orange boxes (labeled with $2$)
multiple lines exist with almost equal gradient and limits.
Careful observation reveals that multiple, almost identical lines exist in each
box.
The ideal situation would be for every line to contain points existing in
a single obstacle surface.
In order to overcome these drawbacks, two more algorithmic stages were implemented;
the breakdown and merging of lines.

The lines breakdown refers to a segmentation procedure, regarding lines
whose points lie in multiple obstacle surfaces.

\begin{figure}
  \centering
  \includegraphics[width=0.9\linewidth]{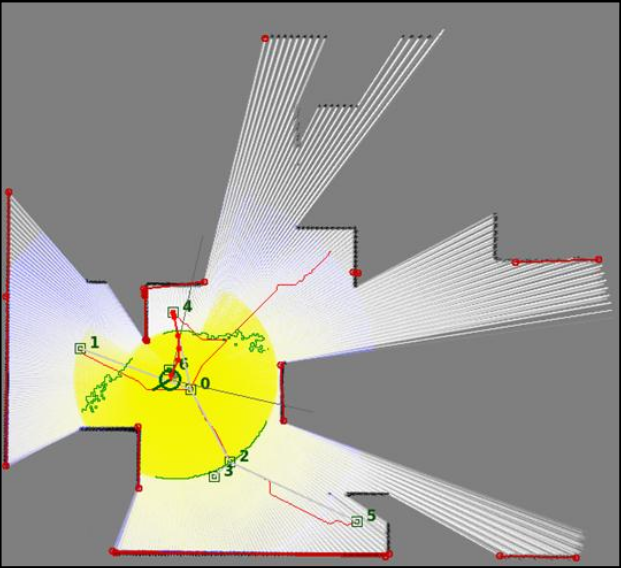}
  \caption{Line segments of the environment after the breakdown procedure}
  \label{fig:ransac2}
\end{figure}

Let's assume that a check is performed in a random line $L_i$.
For each line point $p_j \in L_i$, its distance from the first line
limit $d_j$ is calculated and stored.
Next, an ascending sort is performed in the distances.
This procedure results in sorted points, the first of which are closer to the first
limit and last the ones closer to the second.
The next step is to traverse the distance vector and check if two
successive distances' values difference is larger than a threshold, i.e. if
$|d_k-d_{k+1}|<D_s$.
If this occurs, it can be safely assumed that the specific line contains elements
on different obstacles.
Once the latter is true, a new line is created containing the
elements between the distances' ``gaps'', including the first and last
point.
In figure \ref{fig:ransac2}, the algorithmic result is depicted,
assigning $D_s = 30~px$.

\begin{figure}[htbp]
  \centering
  \includegraphics[width=\linewidth]{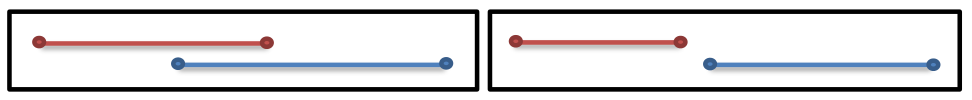}
  \caption{Line segments' merging cases}
  \label{fig:ransac_lines_merge}
\end{figure}

This result indicates that the first problem risen, i.e. the existence of lines
residing in multiple obstacles, is eliminated.
Nevertheless, the second problem remains, as there are still obstacle
surfaces represented by multiple lines with identical characteristics.
For this reason, a line merging algorithm is necessary.

The algorithm that detects similar line segments and merges them is initiated
by checking all lines by pairs.
It is assumed that lines $L_i, L_j$ are checked and their gradients
are $\theta_i, \theta_j$.
Initially, the condition $|\theta_i-\theta_j|<T_{\theta}$ is checked, where
$T_{\theta}$ is the angle threshold for two segments to be considered aligned.
If this condition is met, the proximity of these lines is investigated.
There are two distinct cases to be investigated, presented in
figure \ref{fig:ransac_lines_merge}.

\begin{figure}
  \centering
  \includegraphics[width=.953\linewidth]{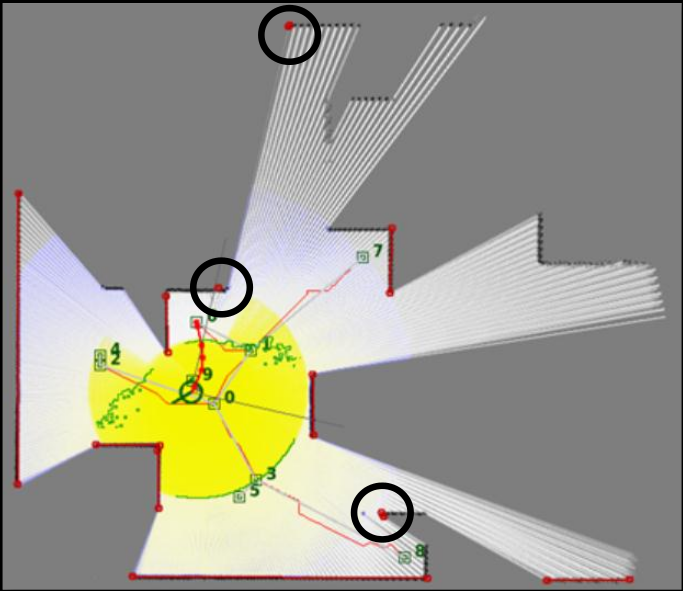}
  \caption{Line segments after the merging procedure}
  \label{fig:ransac3}
\end{figure}

In any of these cases, we check if the limit of the second segment
nearest to the first, has a distance from it lower than a threshold $T_j$.
If this is true, the two lines can be merged, as they are close to each
other and almost parallel.
The merging result is presented in figure \ref{fig:ransac3}.

It is obvious that after the merging technique each
obstacle is represented by a single line.
An additional problem is that the resulting lines contain some
erroneous ones,  comprising only a few points (depicted in
black circles).
These result from the breakdown operations and need to be eliminated,
as their gradient can be random.
Additionally, a case exists (not depicted in figure \ref{fig:ransac3}),
where a line is not entirely aligned to an obstacle surface.
Since our objective is to create an accurate representation of the obstacles,
i.e. line segments with accurate gradients, such misaligned behaviors must not
exist, in order not to participate in the OGM direction vector's calculation.
The first part of the final test a segment must pass, is for its length to be
larger than a minimum threshold, i.e. $||L_i||>T_{Len}$.
In our implementation this was equal to $T_{Len}=30~px$.
The second part is to check each line's reliability.
As \emph{reliability} $L_{Rel}$ we define the mean arithmetic value of the occupancy
probabilities of all line elements.
It is obvious that the ideal case is $L_{rel} \approx 1$, where the line lies
entirely on an obstacle surface.
On the contrary, if $L_{rel}$ is low, this means that the line has parts not
contained in any obstacle surface, thus it should be rejected.
For our case $L_{rel}=0.6$.

\begin{figure}
  \centering
  \includegraphics[width=.9\linewidth]{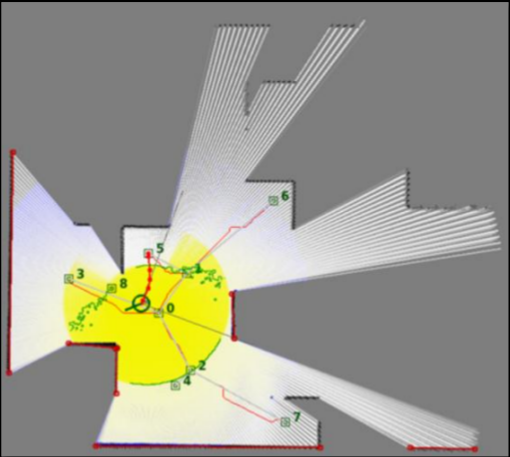}
  \caption{Line detection algorithm's final result}
  \label{fig:ransac_final}
\end{figure}

The final result of the overall method is depicted in figure
\ref{fig:ransac_final}.

\subsubsection{RANSAC Line Segments Quaternion Classification}
\label{ssub:ransac_line_segments_quaternion_classification}

\begin{figure}
  \centering
  \includegraphics[width=.9\linewidth]{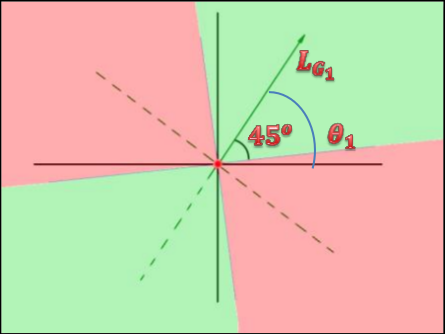}
  \caption{Creation of quadrants based on the first segment}
  \label{fig:gradients}
\end{figure}

The second and final step of the OGM Vector calculation is the segments'
classification in two groups, containing perpendicular lines to each other.
Whilst this is a simple concept, several problems exist regarding its
implementation.
A simple solution would reside to a K-Means algorithm application,
which would group the angles (thus the lines) in two groups.
In reality, this cannot be performed, as segments' gradient angles are
heterogeneous in value.
It should be mentioned that the gradient of a segment to the horizontal axis
is calculated by utilizing the inverse tangent, using the line's extreme
points.
Thus, it is possible for two aligned segments to present opposite gradients
(for example the first to be $89^{\circ}$ and the second $271^{\circ}$).
So, even if these lines are parallel to each other, their
gradient values show opposite directions.
The procedure followed for the segments grouping, is based on the elimination
of such cases and specifically on addition or subtraction of $\pi$ where
necessary.
In order to better understand the algorithm, figure~\ref{fig:gradients} is
presented.

The algorithm takes under assumption that the lines produced by the previous
procedures are indeed perpendicular, i.e. the environment is strongly
rectangular.
The steps followed are:

\begin{enumerate}
  \item{
      The first line is taken under consideration and is denoted as $L_{G_1}$.
      At the same time, four quadrants are created based on the first line's
      gradient $\theta_1$ (figure \ref{fig:gradients}).
    }
  \item{
      The rest of the lines are checked one at a time.
      \begin{itemize}
        \item{
          A segment $L_i$ belongs in the first group ($G_1$) when one of the
          following conditions apply:
          \[
            \begin{array}{ccc}
              \theta_1 + 135^{\circ} \le & \theta_i & \\
                 \theta_1-45^{\circ} \le & \theta_i & \le \theta_1 + 45^{\circ} \\
                                         & \theta_i & \le \theta_1-135^{\circ}
            \end{array}
          \]
          If the segment's gradient lies in the
          green colored quadrants, this line is stored in the first group.
          Additionally, in order for all the parallel segments to the first line
          to acquire a gradient in the first quadrant, we perform:
          \begin{align*}
            \theta_1+135^{\circ}\le\theta_i &\Rightarrow \theta_i = \theta_i-\pi \\
            \theta_i\le\theta_1-135^{\circ} &\Rightarrow \theta_i =\theta_i+\pi
          \end{align*}
        }
        \item{
          The same procedure is applied to the lines that belong to the second
          group ($G_2$), i.e.:
          \[
            \begin{array}{ccc}
              \theta_1+45^{\circ} \le & \theta_i & \le\theta_1+135^{\circ} \\
               \theta_1-45^{\circ}\ge & \theta_i & \geq\theta_1-135^{\circ}
            \end{array}
          \]
          These lines are grouped in the quadrant where the first of them lies.
        }
      \end{itemize}
    }
\end{enumerate}

In order to eliminate any ambiguity concerning the algorithm,
an arithmetic example will be presented.
Let's assume we have identified the lines presented in
table~\ref{tab:line_grouping_example}.

\begin{table}[htpb]
  \centering
  \begin{tabular}{|c|r @{$^\circ$~}||c|r@{$^\circ$~}|}
    \hline
    ID & \multicolumn{1}{c||}{$\theta_i$} & ID & \multicolumn{1}{c|}{$\theta_i$}\\
    \hline
    0 & 2 & 3 & -89   \\ \hline
    1 & 90 & 4 & 179  \\ \hline
    2 & -1 & 5 & 79   \\ \hline
  \end{tabular}
  \caption{Initial lines gradients}
  \label{tab:line_grouping_example}
\end{table}

The steps followed are:

\begin{itemize}
  \item{
      The line with $ID=0$ and $\theta_0=2^{\circ}$ is checked.
      Based on this, the following quadrants are specified: $[-133^{\circ},
      -43^{\circ}], [-43^{\circ}, 47^{\circ}], [47^{\circ}, 137^{\circ}],
      [137^{\circ}, -133^{\circ}]$.
      Obviously, the last quadrant is the merging of the two ranges
      $[137^{\circ}, 180^{\circ}]$ and $[-180^{\circ}, -133^{\circ}]$, since
      we assume that $\theta_i \in \left[ -180^{\circ}, 180^{\circ} \right]$.
      The line with $ID=0$ is assigned to the first group.
    }
  \item{
      The line with $ID=1$ and $\theta_1=90^{\circ}$ is checked.
      Since its gradient is in the range $[47^{\circ}, 137^{\circ}]$, it is
      assigned to the second group.
    }
  \item{
      The line with $ID=2$ and $\theta_2=-1^{\circ}$ is checked.
      Since its gradient lies in the range $[-43^{\circ}, 47^{\circ}]$, it is
      assigned to the first group.
    }
  \item{
      The line with $ID=3$ and $\theta_3=-89^{\circ}$ is checked.
      Since its gradient lies in the range $[-133^{\circ}, -43^{\circ}]$, it is
      assigned to the second group and $\theta_3=\theta_3+\pi=91^{\circ}$.
    }
  \item{
      The line with $ID=4$ and $\theta_4=179^{\circ}$ is checked.
      Since its gradient lies in the range $[137^{\circ}, -133^{\circ}]$, it is
      assigned to the first group and $\theta_4=\theta_4-\pi=-1^{\circ}$
    }
  \item{
      The line with $ID=5$ and $\theta_5=79^{\circ}$ is checked.
      Since its gradient lies in the range $[47^{\circ}, 137^{\circ}]$, it is
      assigned to the second group.
    }
\end{itemize}

The lines produced after the gradient's recalculation are visible in
table~\ref{tab:line_grouping}.

\begin{table}[htbp]
  \centering
  \begin{tabular}{|c|r @{$^\circ$~}||c|r@{$^\circ$~}|}
    \hline
    ID & \multicolumn{1}{c||}{$\theta_i$} & ID & \multicolumn{1}{c|}{$\theta_i$}\\
    \hline
    0 & 2 & 3 & 91   \\ \hline
    1 & 90 & 4 & -1  \\ \hline
    2 & -1 & 5 & 79   \\ \hline
  \end{tabular}
  \caption{Lines gradients after classification}
  \label{tab:line_grouping}
\end{table}

The next step is to calculate the direction of the two vectors defined by the
obstacle line segments' gradients.
It should be reminded that, up to this point, we have accurately calculated the
lines formed by the obstacles and classified them in two alignment groups.

The first group's direction vector is calculated based on the following
weighted mean:

\begin{equation}
  \widetilde{\theta_1}=\frac{ \sum_{\forall L_i \in G_1} \theta_i \cdot ||L_i||\cdot
  (1-L_{Rel}^i)}
  {\sum_{\forall L_i \in G_1} ||L_i||\cdot(1-L_{Rel}^i)}
\end{equation}

This equation sums all the angles' gradients belonging to the
first group ($G_1$), multiplied by a coefficient.
This coefficient is the line's length, multiplied by the one's complement
of the reliability coefficient.
We suppose that a long line is more reliable concerning its gradient, compared to
a short one.
Additionally, if a line is highly reliable, i.e. if $L_{Rel}^i\approx 0$,
it represents quite accurately the surface it lies on.
Conclusively, the direction vector of the first group is calculated via a
weighted mean of this group's lines' gradients, weighted by
coefficients indicative of the lines' ``alignment'' to the obstacles.
Similarly, the second group's direction vector is
calculated (denoted as $\widetilde{\theta_2}$).
The final OGM direction vector is simply
$\theta_{OGM} = \frac{\widetilde{\theta_1}+\widetilde{\theta_2}}{2}$.

\subsubsection{Quadrant Determination via Common RFID Tags}
\label{ssub:quadrant_determination_via_common_rfid_tags}

\begin{figure*}
  \centering
  \includegraphics[width=\linewidth]{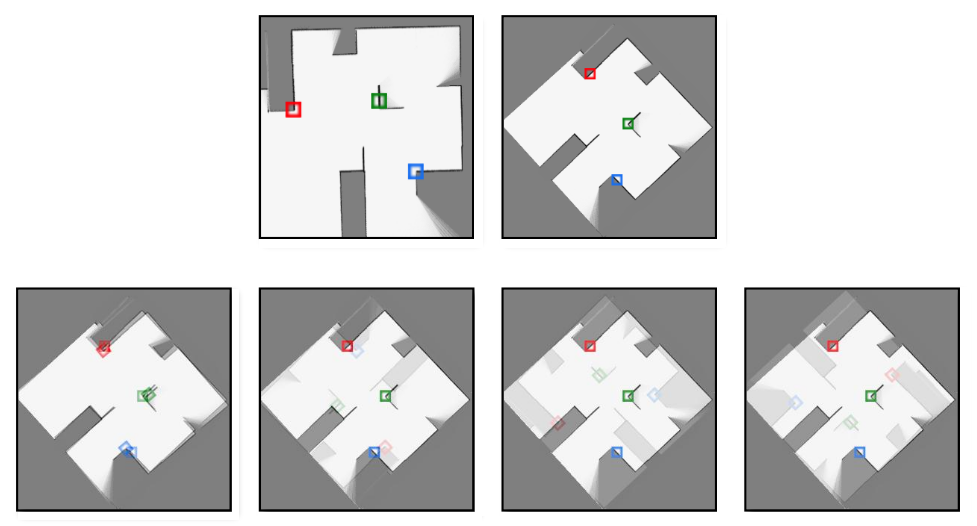}
  \caption{Top: Initial OGMs. Bottom: Four different alignment cases}
  \label{fig:merging_cases}
\end{figure*}

Up to this point, we have determined the OGM's direction vector.
Let's assume that two robotic agents $R_1, R_2$ have reached a point where
their OGMs ($M_1, M_2$) must be merged.
According to the described method, each OGM's direction vector angle,
$\theta_{OGM_1}$, $\theta_{OGM_2}$ is calculated.
Thus, it can be inferred that if the $M_2$ map is rotated by an angle of
$\Delta\theta_{OGM}=\theta_{OGM_1}-\theta_{OGM_2}$,
the two maps' obstacles will be aligned.
Nevertheless, it is understandable, that we have achieved only the alignment and not
their exact match, since this occurs for one of the following rotation
angles:
\[
  \widehat{\Delta\theta_{OGM}} =
  \left\{
  \begin{array}{c}
    \theta_D \\
    \theta_D +\frac{\pi}{2} \\
    \theta_D+\pi \\
    \theta_D+\frac{3\pi}{2}
  \end{array}
\right.
\]

This fact's visualization is depicted in figure~\ref{fig:merging_cases}.
As indicated, the top two images are the initial OGMs ($M_1, M_2$),
in which the three common RFID tags, necessary for the merging initialization,
are marked.
As stated, it is possible to calculate the angle $\Delta\theta_{OGM}$
by which if $M_2$ is rotated, its obstacles will be aligned to the ones of
$M_1$.
It is obvious that only one of the four rotation cases is valid, thus a method
has to be constructed regarding its determination.

It is assumed that each robot has detected three common RFID tags, a
necessary fact for the merging procedure to begin.
The RFID pairs are sorted based on the pair's minimum
localization probability (an example set is depicted in
table~\ref{tab:rfid_tag_merge_example}).

\begin{table}[htbp]
  \centering
  \begin{tabular}{|@{$\,$}c@{$\,$}||@{$\,$}c@{$\,$}|@{$\,$}c@{$\,$}|@{$\,$}c@{$\,$}|@{$\,$}c@{$\,$}|}
    \hline
    \begin{tabular}[c]{@{}c@{}} Tag ID \end{tabular}
    &
    \begin{tabular}[c]{@{}c@{}} Localization\\ probability \\ of robot \#1 \end{tabular}
    &
    \begin{tabular}[c]{@{}c@{}} Localization\\ probability \\ of robot \#2 \end{tabular}
    &
    \begin{tabular}[c]{@{}c@{}}  Minimum\\ common \\precision\end{tabular} &
    \begin{tabular}[c]{@{}c@{}}  Sorting\\ index\end{tabular}
    \\
    \hline
    5 & 82\% & 92\% & 82\% & 2 \\ \hline
    7 & 99\% & 76\% & 76\% & 3 \\ \hline
    9 & 91\% & 92\% & 91\% & 1 \\ \hline
  \end{tabular}
  \caption{Example of three common RFID tags}
  \label{tab:rfid_tag_merge_example}
\end{table}

This way, the most reliable pairs are calculated, which will be used
next.
Let's assume that $t_{i}^j$ is the RFID tag, localized by the robot $i$ with
a sorting index of $j$ and $p_{i}^j$ is its pose in the $M_i$ coordinate
system.
Our known variables are the three pose pairs:
$\left[ p_{1}^1,p_{2}^1 \right],
\left[ p_{1}^2,p_{2}^2 \right],
\left[ p_{1}^3,p_{2}^3 \right]$.
The method followed for the correct quadrants detection, includes the
determination of the lowest probability tag direction, relatively to the
spatial mean of the two other high probability tag poses, in both of the
coordinate systems.

\begin{figure}[h]
  \centering
  \subfigure[First robot map and RFID tag representation]
  {
    \centering
    \includegraphics[width=0.45\linewidth]{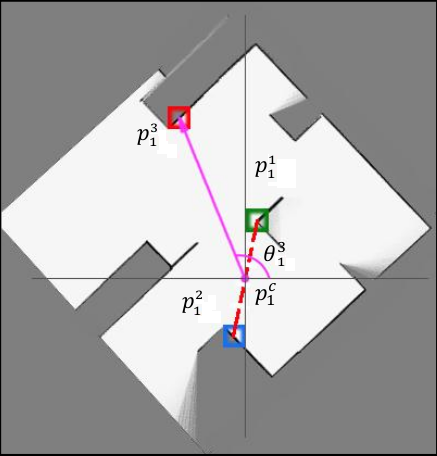}
    \label{fig:quadrant1}
  }
  \hspace{2mm}
  \subfigure[Second robot map and RFID tag representation]
  {
    \centering
  \includegraphics[width=0.45\linewidth]{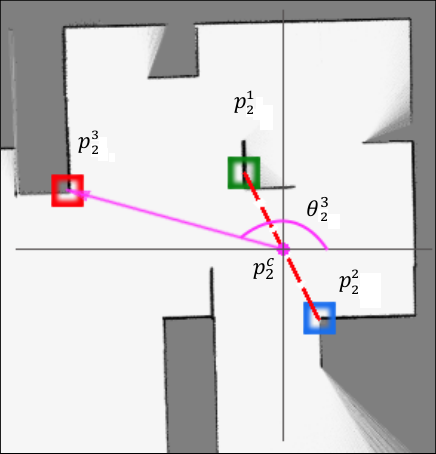}
    \label{fig:quadrant2}
  }

  \caption{OGM rotation quadrant determination}
  \label{fig:quadrant_determination}
\end{figure}

Next, the quadrant occurs by subtracting the two directions.
An example is visible in the figure~\ref{fig:quadrant_determination}.

For each robot's $R_i$ OGM, the following procedure is followed:
Initially, the calculation of the spatial mean of the two most reliable
tags (e.g. $p_{i}^1,p_{i}^2$) is performed:

\begin{equation}
  p_i^c = \left[x_{p_i^c}, y_{p_i^c}\right] = \left[
    \frac{x_{p_i^1} + x_{p_ i^2}}{2},
  \frac{y_{p_i^1} + y_{p_i^2}}{2} \right]
\end{equation}

Next, the angle $\theta_{i}^3$ is calculated, which is the angle of the
line segment defined by the $p_i^c, p_{i}^3$ points.

\begin{equation}
  \theta_{i}^3 = \arctan(y_{p_i^3} - y_{p_i^c},
  x_{p_i^3} - x_{p_i^c})
\end{equation}

Finally, the relative angle between the corresponding line segments in the two
OGMs is $\widehat{\theta_3}=\theta_{1}^3-\theta_{2}^3$.

\begin{figure}
  \centering
  \includegraphics[width=.8\linewidth]{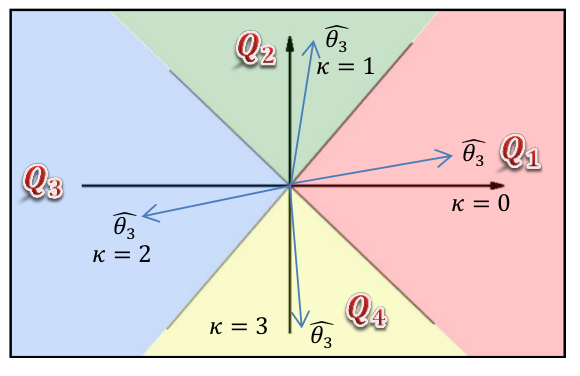}
  \caption{Determination of the correct rotation quadrant}
  \label{fig:quadrant}
\end{figure}

After the computation of the two direction vectors,
the two desired angles $\theta_{OGM_1}$ and $\theta_{OGM_2}$ are available.
The relative rotation angle by which the second robot's map must be rotated,
relatively to the first robot's map is $\Delta\theta_{OGM}=
\theta_{OGM_1}-\theta_{OGM_2}$.
It is understandable that this angle is altered as such:
$\widehat{\Delta\theta_{OGM}}=\theta_{OGM_1}-\theta_{OGM_2}+
\kappa\frac{\pi}{2}$, where $\kappa$ is an integer in the
range $[0,3]$,  defining the angle correction in order not only to align
the maps, but to actually merge them.
The $\kappa$ coefficient is determined as follows (figure~\ref{fig:quadrant}) :

\begin{equation}
  \kappa = \left\{
  \begin{array}{llclc}
    0 & \mbox{if } & \widehat{\theta_3}>-45^{\circ} &
    \mbox{ \& } & \widehat{\theta_3}<45^{\circ} \\
          1 & \mbox{if }&\widehat{\theta_3}>45^{\circ} &
    \mbox{ \& } & \widehat{\theta_3}<135^{\circ} \\
          2 & \mbox{if }&\widehat{\theta_3}>135^{\circ} &
    \mbox{ \& }&\widehat{\theta_3}<215^{\circ} \\
          3 & \mbox{if }&\widehat{\theta_3}>215^{\circ} &
    \mbox{ \& }&\widehat{\theta_3}<305^{\circ} \\
  \end{array}
  \right.
\end{equation}

Conclusively, up to this point, the angle $\widehat{\Delta
\theta_{OGM}}=\theta_{OGM_1}-\theta_{OGM_2}+\kappa\frac{\pi}
{2}$ is calculated, by which $M_1$ must be rotated
relatively to $M_2$ in order for their obstacles to match.
Thus the rotation of the desired transformation has been identified and the only
piece of the puzzle missing is the translation.

\subsection{Relative Translation Calculation via RFID Localization}
\label{sub:translation_calculation_rfid}

The determination of the relative translation between two maps is a quite
easier problem, residing in discovering a single common point in both maps.
This point already exists in our case, and is the RFID tag with the highest
localization probability.

Since each RFID tag is unique, we can directly associate the two maps, by
aligning the two points in the corresponding coordinate systems.
Additionally, as aforementioned, there is a prerequisite that the specific
tag has a localization probability higher than $90\%$.

Conclusively, for the two maps to be able to merge, the following steps must
occur:

\begin{itemize}
  \item{
      Transform $M_2$'s points relatively to the coordinate system
      created by the most reliable RFID tag (i.e. the $p_2^1$).
    }
  \item{
      Rotate these points by $\widehat{\Delta\theta_{OGM}}$ with $p_2^1$
      as center.
    }
  \item{
      Translate the points relatively to the most reliable tag in $M_1$,
      whose coordinates are $\left[x_{p_1^1},y_{p_1^1}\right]$.
    }
\end{itemize}

Thus, for a random point $[x,y]$ existing in the $M_2$ coordinate system,
the transformation procedure into the $M_1$ coordinate system is:

%\begin{align}
  %\left[ \begin{array}{c}
      %x' \\ y'
  %\end{array}\right] =
  %\left[ \begin{array}{cc}
      %cos(\widehat{\Delta\theta_{OGM}}) & -sin(\widehat{\Delta\theta_{OGM}}) \\
      %sin(\widehat{\Delta\theta_{OGM}}) & cos(\widehat{\Delta\theta_{OGM}})
  %\end{array}\right]
  %&
  %\left[ \begin{array}{c}
      %x-x_{p_2^1} \\ y-y_{p_2^1}
  %\end{array}\right] \nonumber \\  + &
  %\left[ \begin{array}{c}
      %x_{p_1^1} \\ y_{p_1^1}
  %\end{array}\right]
%\end{align}

\begin{equation}
  \left[ \begin{array}{c}
      x' \\ y'
  \end{array}\right] = R(\widehat{\Delta\theta_{OGM}})
  \left[ \begin{array}{c}
      x-x_{p_2^1} \\ y-y_{p_2^1}
  \end{array}\right] +
  \left[ \begin{array}{c}
      x_{p_1^1} \\ y_{p_1^1}
  \end{array}\right],
\end{equation}

$R(\theta)$ denotes the two dimensional rotation matrix:

\begin{equation}
  R(\theta) =
  \left[ \begin{array}{cc}
      cos\theta & -sin\theta \\
      sin\theta & cos\theta
  \end{array}\right]
\end{equation}

As it will be shown next, this transformation is not quite exact.
For this reason, we decided to improve the result using an ICP procedure,
as described in the next section.

\subsection{ICP Based Transformation Refining}
\label{sub:icp_transformation}

The final phase of the correct transformation calculation resides in the
execution of an algorithm, aiming to improve the alignment of the two maps.
An ideal algorithm concerning this task is the Iterative Closest Point (ICP),
utilized for the alignment of N-dimensional point sets, by the iterative
approach of the two groups transformation.
The scope is the minimization of the
square differences between the points corresponding to the obstacles in both
maps, by calculating the necessary transformation between them.

This algorithm is widely used in SLAM methods involved with scan matching,
aiming at matching consecutive LRF scans, in pattern matching in the image
manipulation field, or in the computer-based construction of three dimensional
models.
One of its largest drawbacks is its tendency to converge to local minima,
something highly unlikely to occur in our case, as the two sets
are already pre-aligned.
Thus, ICP is utilized to correct possible alignment errors concerning the rotation
angle, as the angle's deviation is $\pm1^{\circ}$, or in the translation
derived by the high certainty RFID tag.
The ICP steps follow:

\begin{enumerate}
  \item{
      Association creation between a point from the first group and a point
      of the second, which are supposed to represent common characteristics.
      Usually, the closest point, or the result of a k-NN (k-Nearest Neighbors)
      algorithm, is selected.
    }
  \item{
      The transformation parameters are calculated, using a mean square error
      function.
    }
  \item{
      The points are transformed based on the previous assumption.
    }
  \item{
      Calculation of the total transformation till now.
    }
  \item{
      Next algorithmic iteration.
    }
\end{enumerate}

\begin{figure}
  \centering
  \includegraphics[width=\linewidth]{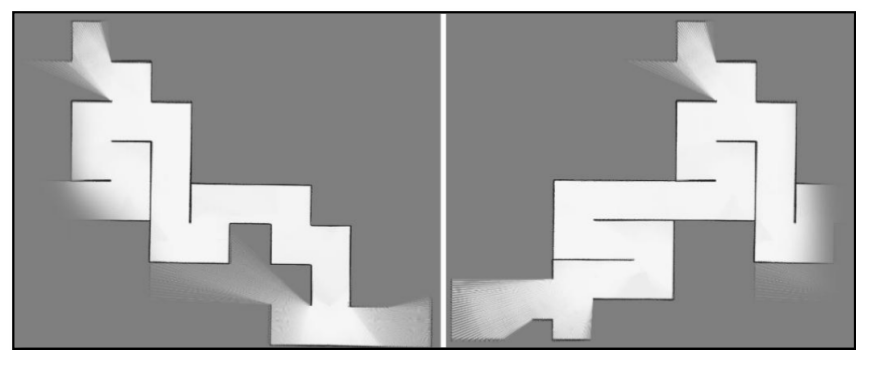}
  \caption{Example of two OGMs for merging}
  \label{fig:convergence_maps}
\end{figure}

Since we desire the best possible alignment of the two maps' obstacles,
the rational criterion of the two groups participating in
the ICP algorithm, is based on the occupied pixels in each map.
This way, the ICP algorithm will try to minimize the mean square error between
them, achieving the best convergence.
Nevertheless, this approach is not globally functional, due
to the mean square error minimization process.
Specifically, in order for the best converge to be achieved, the two sets must
contain elements close to each other.
In order to make this statement understandable, an example of two maps $M_1$
and $M_2$ is depicted in figure~\ref{fig:convergence_maps}.

\begin{figure}
  \centering
  \includegraphics[width=.9\linewidth]{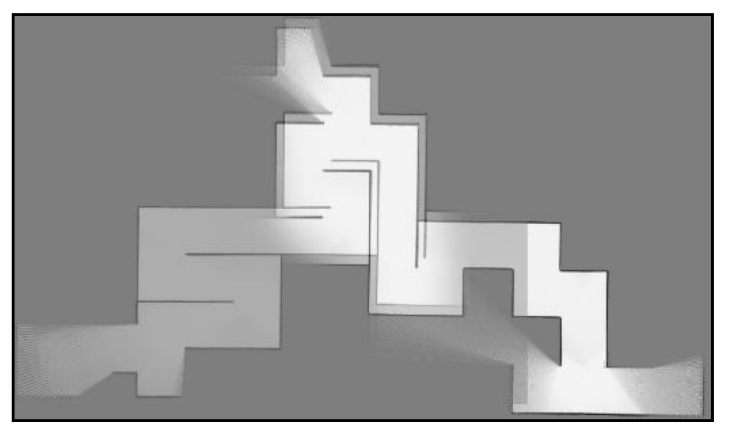}
  \caption{Approximate map merging, based on the initial transformation}
  \label{fig:almost_merged}
\end{figure}

Additionally, let's assume that after the transformation application (based
on the OGM direction vector and the common RFID tags), we got the merging
presented in figure~\ref{fig:almost_merged}.

The merging refinement procedure is obvious to a human being,
but due to the fact that the
two sets corresponding to common areas, contain very few points in
comparison to the entirety of the occupied cells, it is possible for the
ICP to result in total misalignment.
This result is indicative of the ICP behavior to minimize the square sum error,
i.e. to bring as close as possible the points of the two sets.

\begin{figure}
  \centering
  \includegraphics[width=\linewidth]{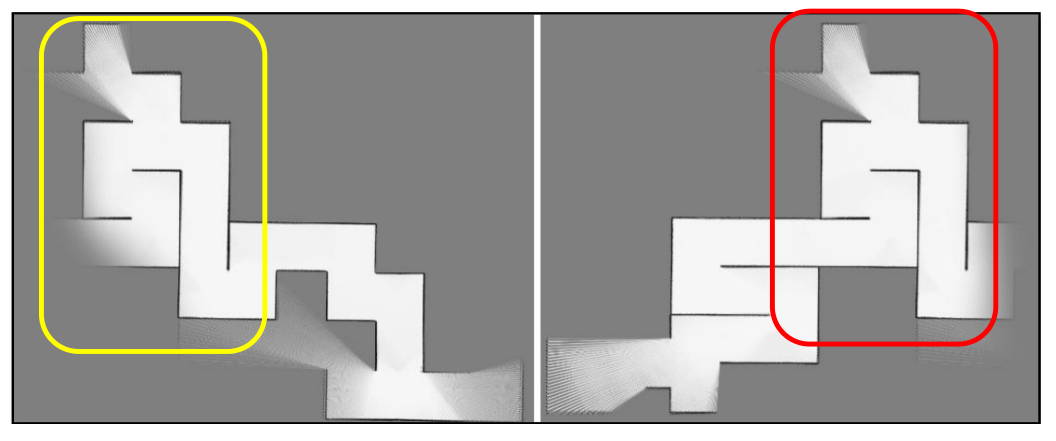}
  \caption{Common section between two OGMs}
  \label{fig:maps_section}
\end{figure}

In order to overcome this problem, specific subsets to
participate in the ICP procedure must be picked, which ought to
present maximum topological similarity.
Thus, the two sets' section in terms of proximity,
after the approximate transformation are selected for employment in the ICP.
This way, the two final sets that will participate in the ICP algorithm, are
the ones shown in the colored rectangles in figure \ref{fig:maps_section}.

We assume that the ICP algorithm terminates with $[\Delta x_{ICP}$, $\Delta
y_{ICP}$, $\Delta \theta_{ICP}]$ as a result.
In conjunction to the previous transformation, the total transformation of
a random point $[x,y]$ from  $M_2$ to $M_1$ coordinate system is performed
by the following procedure:

\begin{enumerate}
  \item{
      $M_2$'s points are relatively translated to the most reliable RFID
      tag, i.e. to $\left[x_{p_2^1},
      y_{p_2^1}\right]$.
    }
  \item{
      These points are rotated by $\widehat{\Delta\theta_{OGM}}$.
    }
  \item{
      The points are translated by $[\Delta x_{ICP}, \Delta y_{ICP}]$.
    }
  \item{
      The points are rotated by $\Delta\theta_{ICP}$.
    }
  \item{
      The points are relatively translated to the $M_1$'s most reliable RFID tag,
      thus relatively to $\left[x_{p_1^1},
      y_{p_1^1}\right]$.
    }
\end{enumerate}

It is obvious that the ICP procedure produces a transformation
intervening the one described in section~\ref{sub:translation_calculation_rfid}.
This fact occurs due to the sets ICP tries to align, which are the
$M_2$ points, after the translation by $\left[x_{p_2^1},
y_{p_2^1}\right]$ and rotation $\widehat{\Delta\theta_{OGM}}$ and
the points of $M_1$ after their translation by $\left[x_{p_1^1},
y_{p_1^1}\right]$.
In mathematical notation, the exact procedure follows:

\begin{equation}
  \label{eq:apply_theta_ogm}
  \left[ \begin{array}{c}
      x' \\ y'
  \end{array}\right] = R(\widehat{\Delta\theta_{OGM}})
  \left[ \begin{array}{c}
      x-x_{p_2^1} \\ y-y_{p_2^1}
  \end{array}\right]
  +
  \left[ \begin{array}{c}
      \Delta x_{ICP} \\ \Delta y_{ICP}
  \end{array}\right]
\end{equation}

\begin{equation}
  \label{eq:apply_theta_icp}
  \left[ \begin{array}{c}
      x'' \\ y''
  \end{array}\right]= R(\Delta\theta_{ICP})
  \left[ \begin{array}{c}
      x' \\ y'
  \end{array}\right]
  +
  \left[ \begin{array}{c}
      x_{p_1^1} \\ y_{p_1^1}
  \end{array}\right]
\end{equation}

The combination of equations~\ref{eq:apply_theta_ogm}~and~\ref{eq:apply_theta_icp},
after applying basic mathematics, result in:

\begin{equation}
  \left[\begin{array}{c} x'' \\ y'' \end{array}\right]
  = R(\Delta\theta_{ICP}+\widehat{\Delta\theta_{OGM}})
  \left[\begin{array}{c} x \\ y \end{array}\right]
  + \left[\begin{array}{c} x_c \\ y_c \end{array}\right]
\end{equation}

We observe that the final transformation formula contains the sum of two
parts.
Whilst the first is relative to the input, the other is constant and
depends only on the two internal transformations parameters.
The constant part is equal to:

\begin{align}
  \left[\begin{array}{c} x_c \\ y_c \end{array}\right] =
  &-R(\Delta\theta_{ICP}+\widehat{\Delta\theta_{OGM}})
  \left[\begin{array}{c}
      x_{p_2^1} \\
      y_{p_2^1}
  \end{array}\right]
  \nonumber \\&+
  R(\Delta\theta_{ICP})
  \left[\begin{array}{c}
      \Delta x_{ICP} \\ \Delta y_{ICP}
  \end{array}\right] +
  \left[ \begin{array}{c}
      x_{p_1^1} \\ y_{p_1^1}
  \end{array}\right]
\end{align}

\subsection{Occupancy Information Merging}
\label{sub:occupancy_information_merging}

Even though the final transformation is accurately calculated, there is still
an additional step required in order to overcome a problem introduced by the
nature of the algorithm itself.
Specifically, since coordinates are integer numbers (specifying cells in a grid),
and should a transformation containing a rotation of arbitrary angle not equal to
$0^{\circ}, 90^{\circ}, 180^{\circ}$ or $270^{\circ}$ be applied,
inconsistencies due to the floating point numbers rounding will occur.

\begin{figure}
  \centering
  \subfigure[Merged OGMs and Coverage fields after a geometrical transformation]
  {
    \includegraphics[width=.9\linewidth]{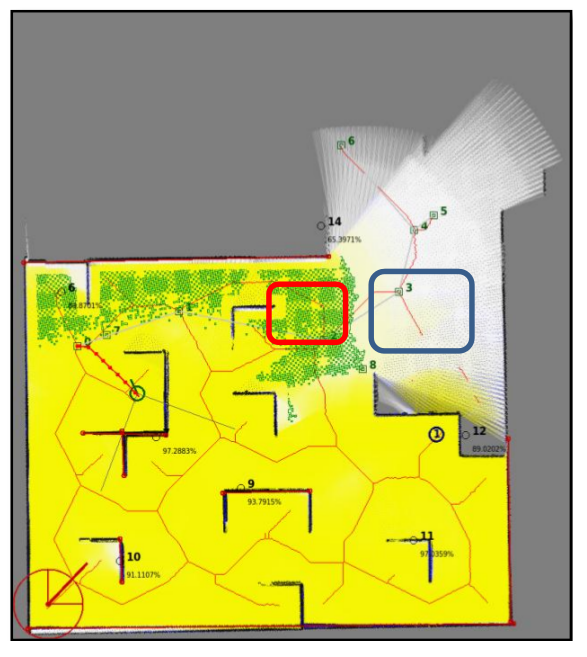}
    \label{fig:ogm_cov_trans}
  }
  \centering
  \subfigure[Inconsistencies in the explored areas]
  {
    \includegraphics[width=0.45\linewidth]{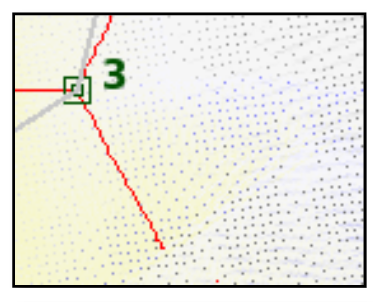}
    \label{fig:trans_zoom}
  }
  \hspace{2mm}
  \centering
  \subfigure[Inconsistencies in the covered areas]
  {
    \includegraphics[width=0.45\linewidth]{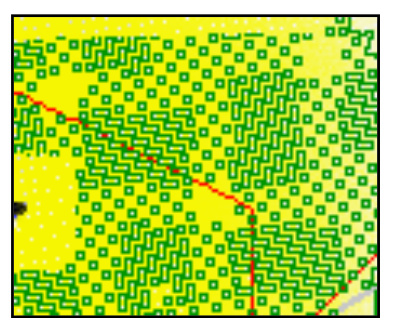}
    \label{fig:trans_zoom_limits}
  }
  \caption{Demonstration of inconsistencies due to merging}
  \label{fig:ogm_merge_incons}
\end{figure}

An illustration of this situation is presented in
figure~\ref{fig:ogm_merge_incons}, whereas in figure~\ref{fig:ogm_cov_trans}
the result of two OGM and coverage fields merging is shown.
Figure \ref{fig:trans_zoom} indicates that after the transformation of
the unoccupied cells of $M_2$ and their merging in $M_1$, the field is not
smooth, but contains unexplored ``gaps''.
The same effect is present in the coverage map
(figure~\ref{fig:trans_zoom_limits}) where inconsistencies appear as well.

\begin{figure}
  \centering
  \includegraphics[width=.9\linewidth]{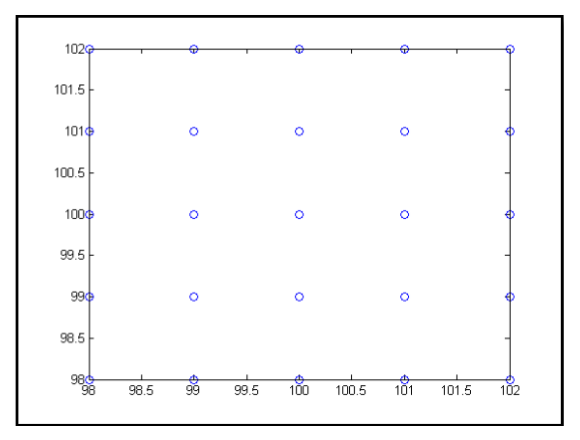}
  \caption{Coordinates field before transformation}
  \label{fig:tr_fi}
\end{figure}

An elaborate example follows, in order to provide an adequate explanation of this
phenomenon.
A field of 25 two dimensional points is assumed, where
$98 \le x \le 102$ and $98 \le y \le 102$, which is rotated by
$30^{\circ}$ around the point $p_{ref}=[200,200]$ and translated in a way
that $p_{ref}$ matches  $p_{ref}'=[300,300]$.
This procedure is a transformation of $dx=100, dy=100, d\theta=30^{\circ}$.
The points before the transformation are depicted in figure~\ref{fig:tr_fi}.

\begin{figure}
  \centering
  \includegraphics[width=.9\linewidth]{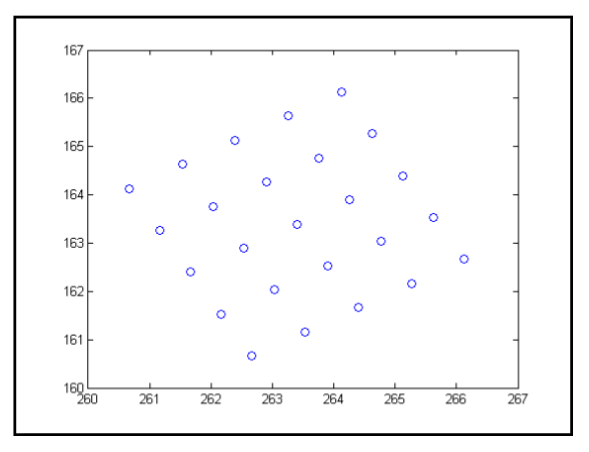}
  \caption{Points field transformation result}
  \label{fig:tr_fi_1}
\end{figure}

\begin{figure}
  \centering
  \includegraphics[width=.9\linewidth]{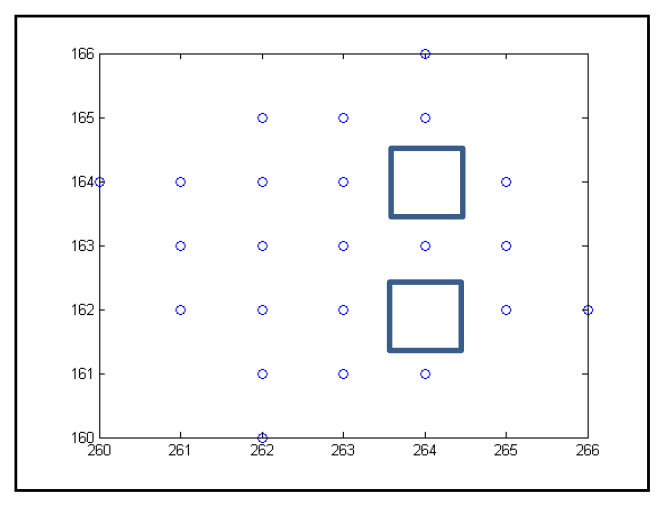}
  \caption{Points field grid transformation result}
  \label{fig:tr_fi_2}
\end{figure}

For a random point $[x,y]$, the applied transformation is:

\begin{equation}
  \left[ \begin{array}{c} x' \\ y' \end{array} \right] =
  R(30^{\circ})
  \left[ \begin{array}{c} x-200 \\ y-200 \end{array} \right] +
  \left[ \begin{array}{c} 300 \\ 300 \end{array} \right]
\end{equation}

This transformation's result is depicted in figure~\ref{fig:tr_fi_1}.
Observing this image, it is inferred that the transformation is correctly
performed (symmetrically and uniformly).
Nevertheless, in the specific figure, the floating point values are depicted.
If this result is translated in a  grid, these values must be rounded,
i.e. become integer.
Thus, the transformation now is:

\begin{equation}
   \left[ \begin{array}{c} x' \\ y' \end{array} \right] =
   \Bigg\lfloor
   R(30^{\circ})
  \left[ \begin{array}{c} x-200 \\ y-200 \end{array} \right] +
  \left[ \begin{array}{c} 300 \\ 300 \end{array} \right]
  \Bigg\rfloor
\end{equation}

The rounding result is evident in figure~\ref{fig:tr_fi_2}.

As expected, the final points are not uniformly distributed, as several
gaps are observed in the grid (blue rectangular shapes).
This fact affects both the merged OGM and the final coverage field.
Concerning the merged OGM, there will be cells left unexplored in the middle
of an explored area, which is undesirable.
Additionally, concerning the coverage fields, the gaps affect the coverage
limits.

Through observation, it is evident that the erroneous cells are presented
either single or in pairs.
Thus, a straightforward approach to smooth the result is to perform a blur mask
(taken from the image processing field) of size $N \times N$.
Since this mask will be applied on the OGM, it is necessary for $N$ to be as
small as possible, or else the produced map's detail will deteriorate.
For this reason $N=3$ was selected, thus the convolution blur mask is equal to:

\begin{equation}
  BlurMask = \frac{1}{9} \cdot
  \left[ \begin{array}{ccc}
      1 & 1 & 1 \\
      1 & 1 & 1 \\
      1 & 1 & 1
  \end{array}\right]
\end{equation}

\begin{figure}[ht]
  \centering
  \subfigure[Merging result, including the blurring technique]
  {
    \includegraphics[width=0.45\linewidth]{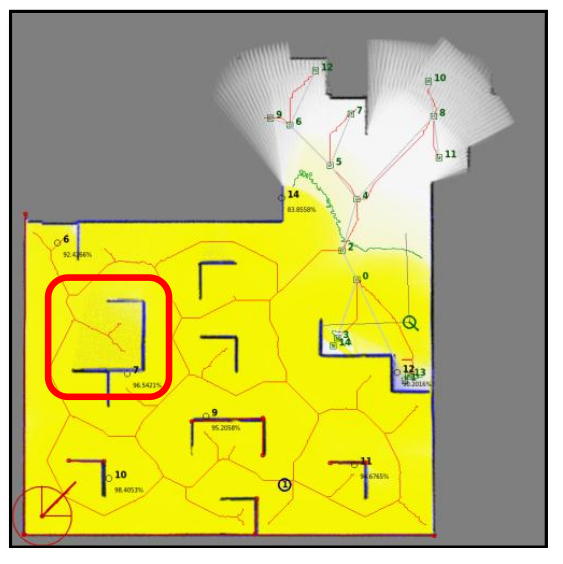}
    \label{fig:blur_ogms}
  }
  \centering
  \subfigure[Detail from the merged and blurred map]
  {
    \includegraphics[width=0.45\linewidth]{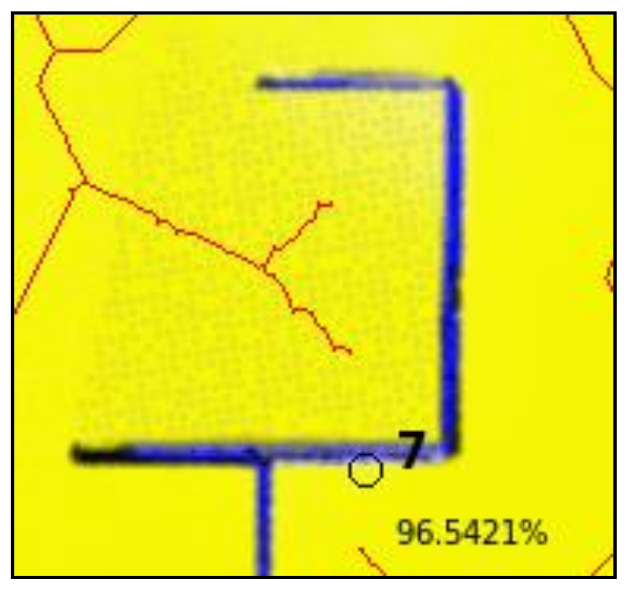}
    \label{fig:blur_ogms_detail}
  }
  \caption{Blurring technique demonstration}
  \label{fig:blur_techn}
\end{figure}

This mask is applied to each OGM and Coverage cell, resulting in the smoothing
of local pixel discontinuities.
Blurring is performed after each merging among a pair of robots.
A merging example, including the blurring technique is presented in
figure~\ref{fig:blur_ogms} and a detail of it in \ref{fig:blur_ogms_detail}.

As observed, the inconsistencies remain to a certain degree, but  are
quite smoothed, overcoming the problems described.

\begin{figure}[ht]
  \centering
  \includegraphics[width=.9\linewidth]{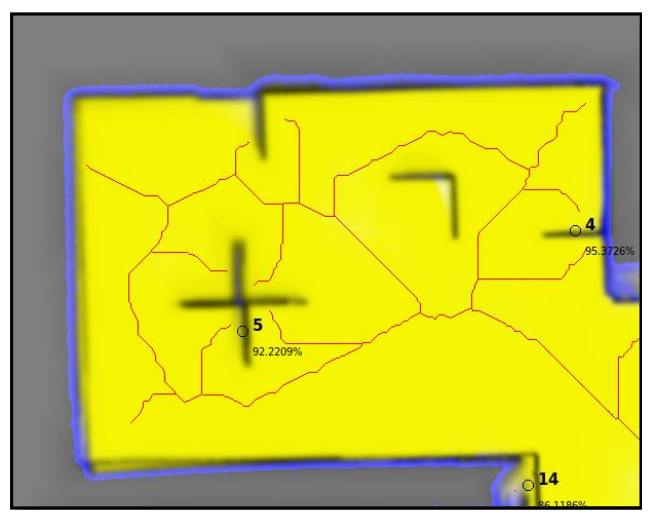}
  \caption{Diffusion effect after a large number of blurring applications}
  \label{fig:blur_diff}
\end{figure}

A problem occurring from the blurring approach, is the loss of the local
information and specifically the large diffusion of the occupied pixels
after multiple merging iterations.
Next, a visualization of this new problem is presented in figure~\ref{fig:blur_diff}.

\begin{figure}[!htb]
  \centering
  \includegraphics[width=.9\linewidth]{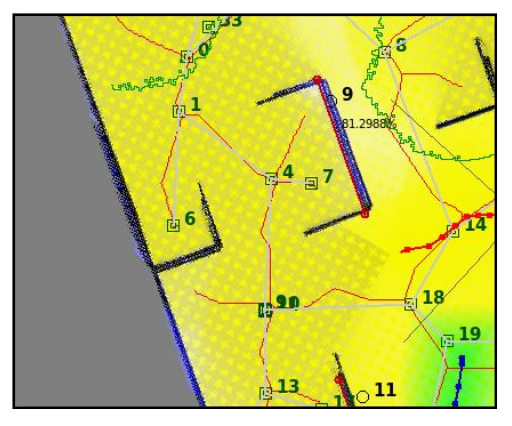}
  \caption{Final merging result}
  \label{fig:final_merging}
\end{figure}

One way to surpass this problem is for the morphological blur kernel to be
applied only under conditions and not in the entirety of the OGM cells.
Specifically, it is necessary to apply it in the explored and unoccupied cells
only, in order not to diffuse occupied values in the free space, or unoccupied
values in the unexplored.
After these assumptions, a sample of the final result is visible in
figure~\ref{fig:final_merging}.

\section{Experiments}
\label{sec:experiments}

In this section, the experiments conducted will be presented.
These are divided in two separate categories, whose combination will allow for
a proper evaluation of both the overall performance and the quality of the results.
Firstly, metrics and results of the OGM Vector calculation will be presented.
Finally, the algorithmic performance of the algorithm in comparison with
other approaches will be investigated.

The platform employed for the experiment execution was
a PC with Intel Core i7 CPU at 2.80GHz, 4GB
RAM, running Ubuntu 9.10 Karmic distribution.

\subsection{OGM Vector Calculation}
\label{sub:ogm_vector_calculation}

\begin{figure}[ht]
  \centering
  \includegraphics[width=.9\linewidth]{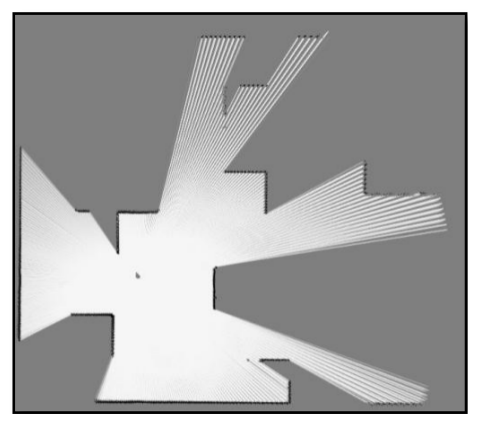}
  \caption{OGM example for direction vector calculation}
  \label{fig:ogm_ex_lines}
\end{figure}

\begin{figure}[ht]
  \centering
  \includegraphics[width=.9\linewidth]{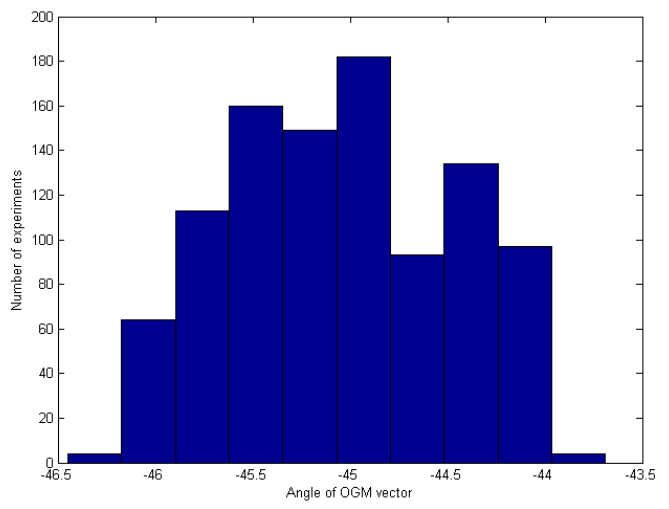}
  \caption{Diagram of the OGM vector angles}
  \label{fig:ogm_vector_angle}
\end{figure}

\begin{figure}[ht]
  \centering
  \includegraphics[width=.9\linewidth]{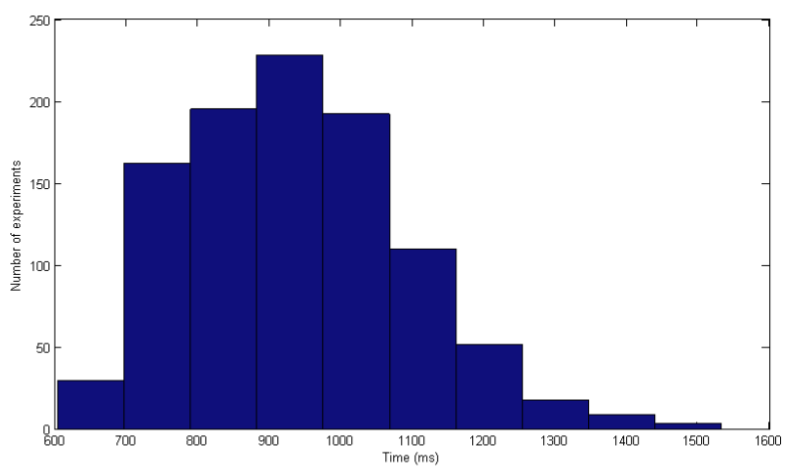}
  \caption{Diagram of the OGM vector execution times in ms}
  \label{fig:ogm_vector_time}
\end{figure}

The environment used (figure~\ref{fig:ogm_ex_lines}) is an OGM, sized $700
\times 600~px$.
Every pixel occupies an area of $0.02
\times 0.02~m^2$, thus the map's dimensions in meters are $14 \times 12~m$.
The OGM direction vector algorithm was executed 1000 times and the metrics are
the computed angle of the OGM vector, as well as the execution time.
The results are depicted in figures~\ref{fig:ogm_vector_angle} and
\ref{fig:ogm_vector_time} respectively.

From the experimental setup, the OGM vector angle is equal to $-45^{\circ}$.
The OGM direction vector's mean angle was $-45.0497^{\circ}$ and the
standard deviation was equal to $0.5598^{\circ}$.
Similarly, regarding the execution times, the mean value was $938.67 ms$ and the
standard deviation $154.05 ms$.

\subsection{Map Merging}
\label{sub:map_merging}

In the current section three map merging methods will be compared:

\begin{enumerate}
  \item{
      Transformation calculation by RFID tags
    }
  \item{
      Transformation calculation by RFID tags and OGM direction vectors
    }
  \item{
      Transformation calculation by RFID tags, OGM direction vectors and
      ICP
    }
\end{enumerate}

The first method, even though not described, is the simplest possible way
to perform transformations by only using the common RFID tags.
Specifically, the transformation computation resides once more in the
calculation of the translation and rotation of one map relatively to another.
The rotation calculation is performed similarly to the section
\ref{ssub:quadrant_determination_via_common_rfid_tags},
i.e. with the formula $\widehat{\theta_3}=\theta_1^3-
\theta_2^3$, where
$\theta_i^3=\arctan (y_{p_i^2}-y_{p_i^c}, x_{p_i^2}-y_{p_i^c})$ and

\begin{equation}
  p_i^c = \left[ x_{p_i^c}, y_{p_i^c} \right]
  = \left[ \frac{x_{p_i^1} + x_{p_i^2}}{2},
  \frac{y_{p_i^1} + y_{p_i^2}}{2} \right]
  \nonumber
\end{equation}

Similarly, the translation computation is performed as such:
Initially, for every tag triplet belonging to a map, their spatial mean
is computed.
For $M_i$, this is:

\begin{equation}
  [x_C^i, y_C^i] = \left[
    \frac{ x_{p_i^1} + x_{p_i^2} + x_{p_i^3} }{3},
    \frac{ y_{p_i^1} + y_{p_i^2} + y_{p_i^3} }{3}
  \right]
\end{equation}

Finally, the transformation is performed by rotating $M_2$ by
$\widehat{\theta_3}$
and translating it by $[x_C^2-x_C^1, y_C^2-y_C^1]$.

The second method resides in the transformation computation, using the rotation
computed by the OGM direction vectors for each map and the translation
occurring by the most accurately localized RFID tag.
Finally, the third method constitutes an extension of the second, as a further
improvement step was added, the ICP algorithm.
For every method, five experiments were conducted and the mean square error of
the first and the transformed second map, as well as the execution time of
each method were measured.

\begin{figure}[h]
  \centering
  \subfigure[Alignment example \#1]
  {
    \centering
    \includegraphics[width=0.75\linewidth]{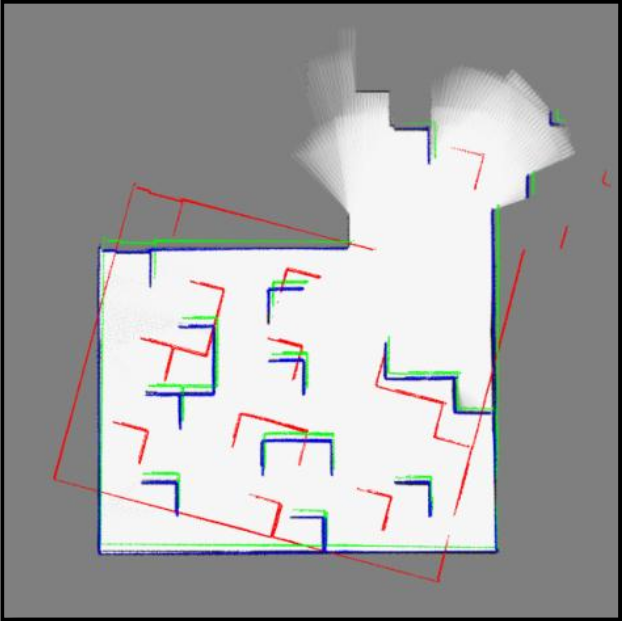}
    \label{fig:exp_align_1}
  }
  \hspace{2mm}
  \subfigure[Alignment example \#2]
  {
    \centering
    \includegraphics[width=0.75\linewidth]{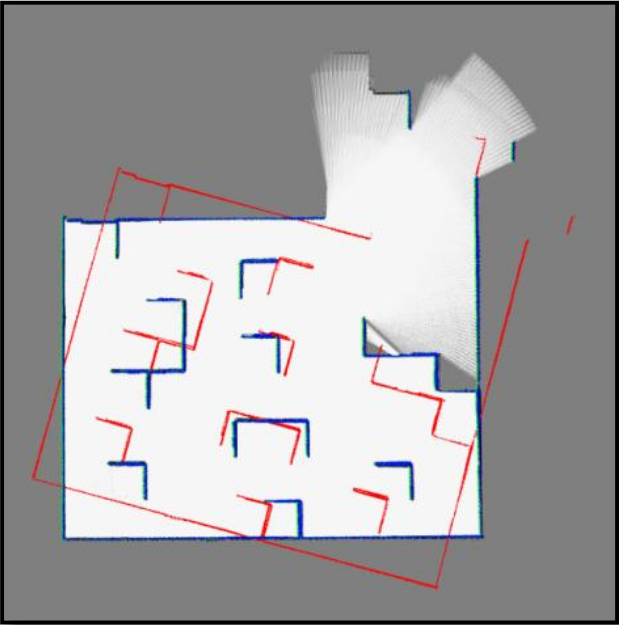}
    \label{fig:exp_align_2}
  }
  \caption{Two alignment examples for all three methods. The first method's
  result is in red color, the second method's in green and the third method's
  in blue.}
  \label{fig:exp_map_align}
\end{figure}

\begin{table}
  \centering
  \begin{tabular}{|@{$\,$}c@{$\,$}||@{$\,$}c@{$\,$}|@{$\,$}c@{$\,$}|@{$\,$}c@{$\,$}|}
    \hline
    \begin{tabular}[c]{@{}c@{}} Mean of five\\experiments \end{tabular} &
    \begin{tabular}[c]{@{}c@{}} Method \#1 \end{tabular} &
    \begin{tabular}[c]{@{}c@{}} Method \#2 \end{tabular} &
    \begin{tabular}[c]{@{}c@{}} Method \#3 \end{tabular}
    \\
    \hline
    \begin{tabular}[c]{@{}c@{}} Mean\\ Square\\ Error \end{tabular} &
    \begin{tabular}[c]{@{}c@{}} 7178.27 $px^2$ \end{tabular} &
    \begin{tabular}[c]{@{}c@{}} 123.45 $px^2$ \end{tabular} &
    \begin{tabular}[c]{@{}c@{}} 2.6 $px^2$ \end{tabular}
    \\
    \hline
    \begin{tabular}[c]{@{}c@{}} Mean\\ Execution\\ Time \end{tabular} &
    \begin{tabular}[c]{@{}c@{}} 0.026 ms \end{tabular} &
    \begin{tabular}[c]{@{}c@{}} 10490.8 ms \end{tabular} &
    \begin{tabular}[c]{@{}c@{}} 18497.6 ms \end{tabular}
    \\
    \hline
  \end{tabular}
  \caption{Map merging experiments}
  \label{tab:map_merging_experiments}
\end{table}

Figure~\ref{fig:exp_map_align} indicates that the first method
produces erroneous transformations that diverge a lot from the desired
ones.
Additionally, the second method results in satisfying transformations, since
it manages to align the obstacles.
Finally, the third method calculates the correct transformation,
since the transformed points of the first map are exactly matched to the
second.

\section{Conclusions \& Future Work}
\label{sec:conclusions}

Initially, as far as the OGM Vector calculation is concerned,
the experimental results showcase the algorithmic precision, since the
angle's mean value is extremely close to the real one.
Additionally, the standard deviation is quite low, from the value of which
we can assume that the $95.45\%$ of the cases have a deviation of $\pm2\cdot
\sigma=\pm 1.1196^{\circ}$, which is quite satisfactory.
Regarding the execution time, its value is not low (almost a second),
but by no means it is prohibitive to our application.

Regarding the map merging algorithm,
the optical results inferred from figure~\ref{fig:exp_map_align} are
strengthened from the mean square errors (MSE)
measured, shown in table~\ref{tab:map_merging_experiments}.
It is obvious, that the first method totally fails in aligning the obstacles correctly,
resulting in a large MSE, the second method has a lower error value and
the third almost zero.
Of course, similarly to any computational procedure, there is a counterweight
regarding precision, which of course is the performance.
Indeed, the experimental results indicate that the more precise the method,
the larger execution time it requires.
Specifically, the worst qualitative method demands the smallest execution
time (a mean of 0.02 ms), since it is the simplest.
The second method introduces the utilization of the RANSAC algorithm, thus
the execution time increased a lot (almost 10 sec), whilst the third inserts
more computations via the ICP algorithm, reaching an almost perfect aligning
roughly in 18 seconds.
Obviously the latter execution time is far from optimal for an on-line
multi robot exploration system, but it is not prohibiting to the overall cause,
since this algorithm is executed at most $N-1$ times in a $N$-agent system.
Conclusively, it is preferable to spend execution resources than having a
misalignment in the map merging procedure.
Concerning the ICP performance issues, it would also been possible to perform
the ICP algorithm, right after the transformation determination deriving only by
the common RFID tags, aiming at eliminating the RANSAC procedure.
This was purposefully not selected, as ICP is a very time demanding algorithm,
fact supported by the experimental results, where nearly 8 seconds
were spend to align almost similar points sets.

As far as future work is concerned, a similar approach can be implemented by
substituting the RFID tags with other high level topological features such as
corridors, doors or even with low level features such as straight line segments
or corners.
This way, no necessity for external devices exists, though the feature
correspondence problem must be addressed.
If one decides to follow the RFID approach, investigation should be performed
regarding the employment of two instead of three common RFID tags, in order
for the merging procedure to initiate.
Even though the merging algorithm initiation conditions will be met in a shorter
time period, enabling the robots to cooperate at an early exploration stage,
further steps should be incorporated, as the alignment of the two local maps
is now topologically ambiguous.
Finally, one of the problems not addressed in this work, is the realistic RFID
tag localization, as certain difficulties are introduced due to the signal
cutoff or reflections.
Consequently, it would be interesting to research our proposal robustness in
realistic, thus more noisy, environments.





\end{document}